\documentclass[conference]{IEEEtran}
\IEEEoverridecommandlockouts
\usepackage{cite}
\usepackage{amsmath,amssymb,amsfonts}
\usepackage{algorithmic}
\usepackage{graphicx}
\usepackage{textcomp}
\usepackage{xcolor}
\usepackage{svg}
\usepackage{balance}
\usepackage{hyperref}

\def\BibTeX{{\rm B\kern-.05em{\sc i\kern-.025em b}\kern-.08em
    T\kern-.1667em\lower.7ex\hbox{E}\kern-.125emX}}
\begin{document}

\title{LLM-Powered Interactive Robotic Action Synthesis from Multimodal Speech, Gestures, and Music}

\author{\IEEEauthorblockN{Snehasis Banerjee, Ranjan Dasgupta}
\IEEEauthorblockA{\textit{Physical AI Research} \\
\textit{TCS Research, Tata Consultancy Services}\\
Kolkata, India \\
\{snehasis.banerjee,ranjan.dasgupta\}@tcs.com}
}

\maketitle

\begin{abstract}
The quest for intuitive and natural human-robot interaction (HRI) remains a significant challenge in robotics. Traditional methods often rely on rigid, pre-programmed commands that limit the robot's expressiveness and adaptability. This paper introduces a novel framework that leverages the reasoning capabilities of Large Language Models (LLMs) to synthesize complex robotic actions from a rich tapestry of multimodal human inputs: natural speech, hand gestures, and music/sound beats. Our system architecture integrates a speech transcription model, a gesture recognition module, and a signal processing pipeline for beat detection. These processed inputs are contextualized using prompt templates and fed into a LLM. The LLM, informed by a predefined robot action space, reasons over the combined inputs to generate a coherent sequence of actions. This sequence is dispatched to an action queue for execution on a quadruped robot over ROS. The framework has ability to interpret and fuse semantic commands from speech, deictic information from gestures, and rhythmic cues from music. This work represents a step towards creating robots that can interact with humans in a more fluid, creative, and context-aware manner.
\end{abstract}



The ability of robots to seamlessly collaborate with and act upon instructions from humans is a cornerstone of next-generation automation. While significant strides have been made in isolated command modalities like speech~\cite{deuerlein2021human} or gesture control~\cite{wang2022hand}, these interactions often lack the richness and contextual nuance of human communication. Humans naturally combine speech, gestures, and even respond to environmental cues like music to convey intent. Enabling robots to understand and act upon this multimodal stream of information is crucial for achieving truly symbiotic HRI.

Recent advancements in Large Language Models (LLMs) have demonstrated unprecedented capabilities in natural language understanding, reasoning, and planning~\cite{kim2024survey}. This has inspired a new wave of research in robotics, where LLMs are used as a central `brain' to translate high-level human goals into low-level robot actions. However, most existing work focuses primarily on text-based inputs, neglecting the valuable information contained in non-verbal modalities.

\begin{figure*}[htbp]
\centering
\includegraphics[width=0.7\linewidth]{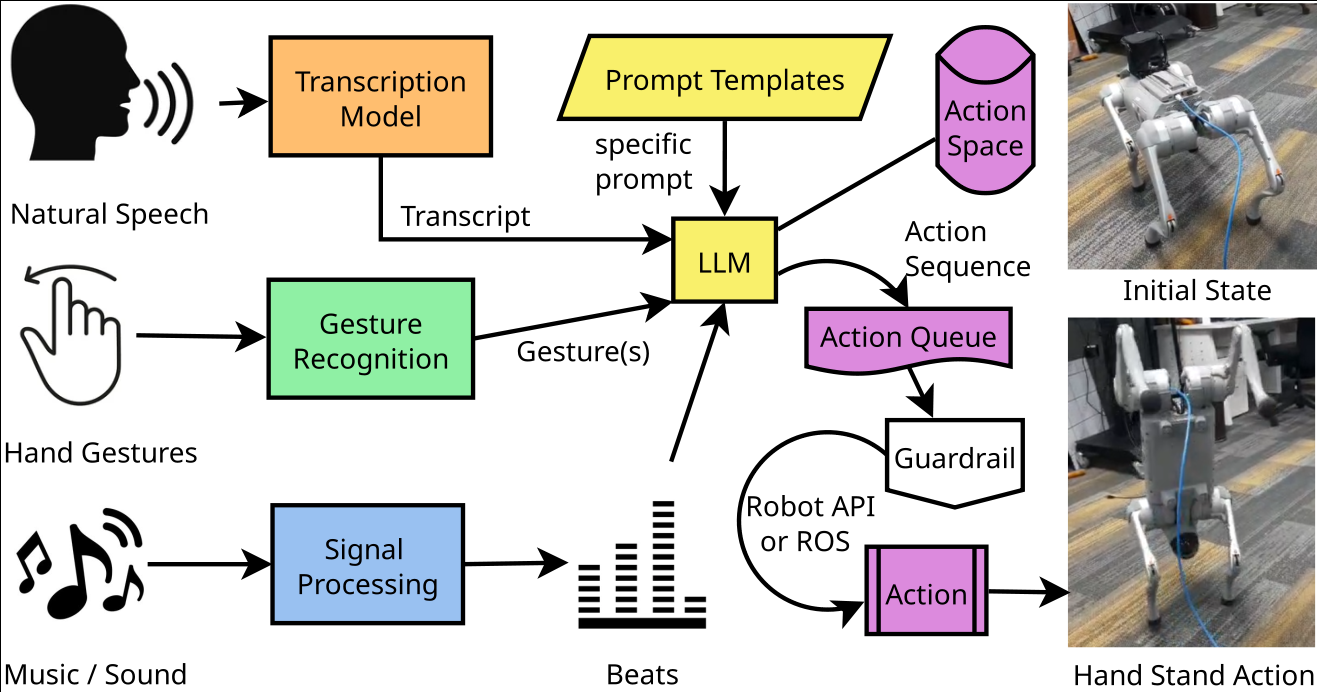}
\caption{System architecture for LLM-powered multimodal action synthesis. Natural speech, hand gestures, and music are processed into structured inputs (transcripts, recognized gestures, and beats) for the LLM, to generate an action sequence for the robot. Example is shown for command `stand on your hand'.}
\label{fig:architecture}
\end{figure*}

This paper presents a framework, illustrated in Figure \ref{fig:architecture}, for multimodal robotic action synthesis. Our key contribution is the integration of three distinct input channels—natural speech, hand gestures, and music beats — into a unified system orchestrated by an LLM with tailored prompts. We hypothesize that this fusion allows for more expressive and sophisticated robot control. For example, a user can say, `Do this trick', while performing a specific gesture, and add, `on the beat', prompting the robot to synthesize an action that is not only semantically correct but also rhythmically synchronized. We validate our approach on a quadruped robot, demonstrating its ability to perform complex maneuvers based on these multimodal commands. Additionally, for near real time system response, the model specifics have been carefully chosen post ablation, that are capable of running in low compute. 

\section{System Architecture}
The proposed framework consists of three main stages: multimodal input processing, LLM-based action reasoning, and robotic execution, as depicted in Figure \ref{fig:architecture}.

\subsection{Multimodal Input Processing}
\begin{itemize}
    \item \textbf{Natural Speech}: User's vocal commands are captured by a microphone and fed into an automatic speech recognition (ASR) model (Whisper~\cite{andreyev2025quantization}). The model outputs a text transcript, which serves as the primary semantic instruction for the LLM.
    \item \textbf{Hand Gestures}: A camera captures the user's hand movements. A real-time gesture recognition module (~\cite{kapitanov2024hagrid}) processes the video stream to identify and classify static or dynamic gestures. The recognized gesture (e.g., 'open palm', 'thumbs up') provides complementary, often deictic information.
    \item \textbf{Music/Sound}: An audio input stream is processed to extract rhythmic information. A signal processing pipeline applies beat detection algorithms~\cite{foscarin2024beat} to identify the tempo and timing of beats in ambient music or sound. This provides a temporal structure for the robot's actions.
\end{itemize}

\subsection{LLM-based Action Reasoning}
The processed inputs from all three modalities are aggregated and formatted into a structured prompt using a set of predefined templates, before sending to the LLM. The LLM's task is to act as a translator and planner. It takes the combined command (e.g., Transcript: `Do a flip', Gesture: `pointing up', Beats: [20 BPM]) and maps it to a plausible sequence of executable actions defined in the robot's `Action Space'. The Action Space is a database or a structured description of all primitive skills the robot can perform (e.g., stand, sit, jump, forward, rotate). The LLM reasons about the goal (user instruction) and constraints (obstacles detected by sensors) to compose these primitives into a high-level plan.

\subsection{Robotic Execution}
The action sequence generated by the LLM is sent to an `Action Queue'. Each action in the queue is translated into a specific API command \footnote{Unitree Go2 SDK - \url{https://github.com/unitreerobotics/unitree_sdk2}} and sent to the robot's controller, via ROS or OMG-DDS\footnote{Implementation as CycloneDDS: \url{https://github.com/eclipse-cyclonedds}}. The robot then transitions from its initial state to execute the commanded action, such as the `Hand Stand Action' shown in the figure.

\section{Implementation and Results}
Our experimental setup utilizes a Unitree Go2 quadruped robot. For speech processing, we employed OpenAI's Whisper `small' model for transcription. Due to near real-time demands,  Qwen3:0.6b~\cite{yang2025qwen3} was used as Ollama-based LLM. In a preliminary experiment, a user instructed the robot, `Do something cool like standing on your limbs when my fist opens' while making a `fist' gesture. Upon the user opening their hand (an implicit `go' gesture) and the start of a beat, the system successfully generated and executed the sequence for a handstand, that lasted till the duration of beat. This demonstrates the LLM's ability to fuse the semantic meaning (`stand on limbs'), the triggering event (gesture change), and the rhythmic constraint (the beat) into a single, coherent robotic action. To mitigate risks from LLM hallucinations, our framework employs a multi-layered safety system. The most critical `Guardrail' is a constrained Action Space; the LLM can only select from a pre-vetted library of safe, high-level skills and cannot generate raw motor commands. Additionally, structured prompts guide the LLM to refuse ambiguous requests. Finally, a deterministic `Action Sequence Validator' verifies any generated plan for correctness before it is sent for execution. This multi-layered approach ensures the robot operates within a predictable and safe envelope, grounding the LLM's generative capabilities in physical reality.

\section{Conclusion and Future Work}
We presented a novel LLM-powered framework for interactive robotic action synthesis that integrates speech, gestures, and music beats. Our initial results show that this multimodal approach enables more expressive and context-aware human-robot interaction. By allowing an LLM to reason over this rich input, robots can perform complex actions that are aligned with a user's multifaceted intent. Future work will focus on expanding the action space, improving the real-time responsiveness of the system by optimizing the LLM inference pipeline, and incorporating a feedback loop where the robot's state can inform the LLM's subsequent planning steps.

\balance

\bibliographystyle{abbrv}
\bibliography{ref}

@article{kim2024survey,
  title={A survey on integration of large language models with intelligent robots},
  author={Kim, Yeseung and Kim, Dohyun and Choi, Jieun and Park, Jisang and Oh, Nayoung and Park, Daehyung},
  journal={Intelligent Service Robotics},
  volume={17},
  number={5},
  pages={1091--1107},
  year={2024},
  publisher={Springer}
}

@article{deuerlein2021human,
  title={Human-robot-interaction using cloud-based speech recognition systems},
  author={Deuerlein, Christian and Langer, Moritz and Se{\ss}ner, Julian and He{\ss}, Peter and Franke, J{\"o}rg},
  journal={Procedia Cirp},
  volume={97},
  pages={130--135},
  year={2021},
  publisher={Elsevier}
}

@inproceedings{wang2022hand,
  title={Hand and arm gesture-based human-robot interaction: a review},
  author={Wang, Xihao and Shen, Hao and Yu, Hui and Guo, Jielong and Wei, Xian},
  booktitle={Proceedings of the 6th International Conference on Algorithms, Computing and Systems},
  pages={1--7},
  year={2022}
}

@article{andreyev2025quantization,
  title={Quantization for OpenAI's Whisper Models: A Comparative Analysis},
  author={Andreyev, Allison},
  journal={arXiv preprint arXiv:2503.09905},
  year={2025}
}

@inproceedings{kapitanov2024hagrid,
  title={HaGRID--HAnd Gesture Recognition Image Dataset},
  author={Kapitanov, Alexander and Kvanchiani, Karina and Nagaev, Alexander and Kraynov, Roman and Makhliarchuk, Andrei},
  booktitle={Proceedings of the IEEE/CVF Winter Conference on Applications of Computer Vision},
  pages={4572--4581},
  year={2024}
}

@article{foscarin2024beat,
  title={Beat this! Accurate beat tracking without DBN postprocessing},
  author={Foscarin, Francesco and Schl{\"u}ter, Jan and Widmer, Gerhard},
  journal={arXiv preprint arXiv:2407.21658},
  year={2024}
}

@article{yang2025qwen3,
  title={Qwen3 technical report},
  author={Yang, An and Li, Anfeng and Yang, Baosong and Zhang, Beichen and Hui, Binyuan and Zheng, Bo and Yu, Bowen and Gao, Chang and Huang, Chengen and Lv, Chenxu and others},
  journal={arXiv preprint arXiv:2505.09388},
  year={2025}
}

\end{document}